\documentclass[sigconf]{acmart}
\AtBeginDocument{%
  }

\setcopyright{none}
\renewcommand\footnotetextcopyrightpermission[1]{}
\copyrightyear{2026}
\acmYear{2026}
\acmDOI{XXXXXXX.XXXXXXX}

\usepackage{amsmath}

\usepackage{amssymb}
\usepackage{booktabs}
\usepackage{multirow}
\usepackage[table]{xcolor}
\usepackage{subcaption}
\usepackage{tcolorbox}
\usepackage{soul}
\usepackage{enumitem}

\tcbset{
    findingbox/.style={
        colframe=black,
        colback=gray!5,
        boxrule=0.5mm,
        arc=4pt,
        auto outer arc,
        left=2mm,
        right=2mm,
        top=1mm,
        bottom=1mm
    }
}

\definecolor{resultblue}{RGB}{235,245,255}
\definecolor{rowblue}{RGB}{230,245,255}

\newcommand{\RNum}[1]{\uppercase\expandafter{\romannumeral #1\relax}}

\begin{document}
\title{SAM3-LiteText: An Anatomical Study of the SAM3 Text Encoder for Efficient Vision-Language Segmentation}

\author{Chengxi Zeng}
\authornote{Corresponding Author.}
\affiliation{%
  \institution{Visual Information Lab, University of Bristol}
  \city{Bristol}
  \country{United Kingdom}
}
\email{simon.zeng@bristol.ac.uk}

\author{Yuxuan Jiang}
\affiliation{%
  \institution{Visual Information Lab, University of Bristol}
  \city{Bristol}
  \country{United Kingdom}
}
\email{yuxuan.jiang@bristol.ac.uk}

\author{Ge Gao}
\affiliation{%
  \institution{Visual Information Lab, University of Bristol}
  \city{Bristol}
  \country{United Kingdom}
}
\email{ge1.gao@bristol.ac.uk}

\author{Shuai Wang}
\affiliation{%
  \institution{University of Amsterdam}
  \city{Amsterdam}
  \country{Netherlands}
}
\email{s.wang3@uva.nl}

\author{Duolikun Danier}
\affiliation{%
  \institution{School of Informatics, University of Edinburgh}
  \city{Edinburgh}
  \country{United Kingdom}
}
\email{duolikun.danier@ed.ac.uk}

\author{Bin Zhu}
\affiliation{%
  \institution{Singapore Management University}
  \city{Singapore}
  \country{Singapore}
}
\email{binzhu@smu.edu.sg}

\author{Stevan Rudinac}
\affiliation{%
  \institution{University of Amsterdam}
  \city{Amsterdam}
  \country{Netherlands}
}
\email{s.rudinac@uva.nl}

\author{David Bull}
\affiliation{%
  \institution{Visual Information Lab, University of Bristol}
  \city{Bristol}
  \country{United Kingdom}
}
\email{Dave.Bull@bristol.ac.uk}

\author{Fan Zhang}
\authornote{Project PI.}
\affiliation{%
  \institution{Visual Information Lab, University of Bristol}
  \city{Bristol}
  \country{United Kingdom}
}
\email{aaron.zhang@bristol.ac.uk}

\renewcommand{\shortauthors}{Zeng et al.}

\begin{abstract}
Vision-language segmentation models such as SAM3 enable flexible, prompt-driven visual grounding, but inherit large, general-purpose text encoders originally designed for open-ended language understanding. In practice, segmentation prompts are short, structured, and semantically constrained, leading to substantial over-provisioning in text encoder capacity and persistent computational and memory overhead. In this paper,
we perform a large-scale anatomical analysis of text prompting in vision–language segmentation, covering 404,796 real prompts across multiple benchmarks. Our analysis reveals severe redundancy: most context windows are underutilized, vocabulary usage is highly sparse, and text embeddings lie on a low-dimensional manifold despite high-dimensional representations.
Motivated by these findings, we propose SAM3-LiteText, a lightweight text encoding framework that replaces the original SAM3 text encoder with a compact MobileCLIP student that is optimized by knowledge distillation.
Extensive experiments on image and video segmentation benchmarks show that SAM3-LiteText reduces text encoder parameters by up to 88\%, substantially reducing static memory footprint, while maintaining segmentation performance comparable to the original model.
Code: \textbf{\url{https://github.com/SimonZeng7108/efficientsam3/tree/sam3_litetext}}.
\end{abstract}



\begin{teaserfigure}
  \centering
  \includegraphics[width=0.9\textwidth]{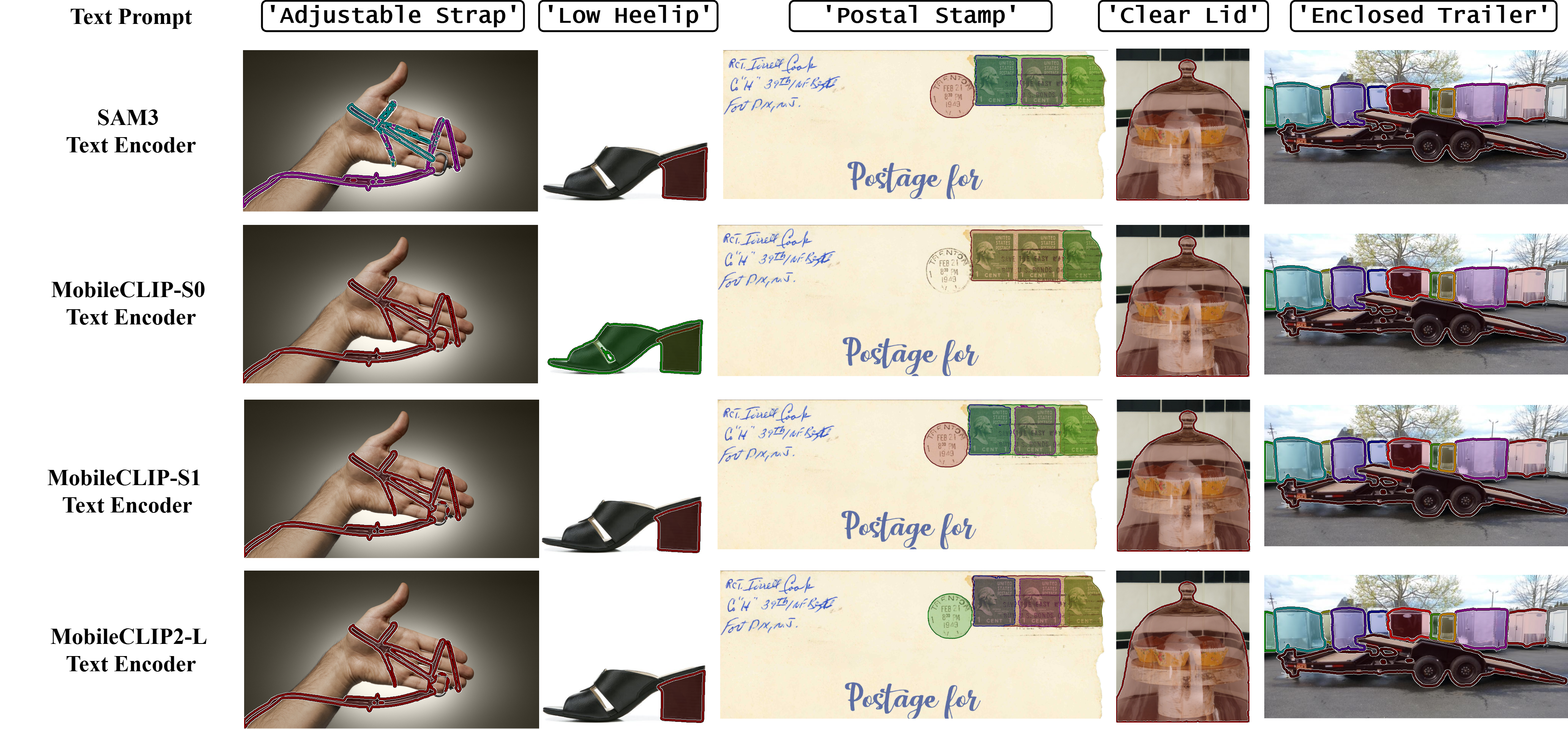}
  \caption{ The top row displays results from the SAM3 teacher model, while subsequent rows show our SAM3-LiteText student variants using MobileCLIP-S0, S1, and MobileCLIP2-L. Despite the significant reduction in model size, our student models (using $L=16$ embeddings) produce segmentation masks that are visually indistinguishable from the teacher's. This demonstrates that our models are drastically smaller yet perform competitively with the original high-parameter teacher model.}
  \label{fig:qualitative}
  \Description{A grid of images comparing segmentation masks between the SAM3 teacher (top row) and our smaller MobileCLIP student models (lower rows). The masks are visually identical despite the student models' smaller size.}
\end{teaserfigure}
\maketitle
\section{Introduction}
\label{sec:intro}

Vision-language foundation models are rapidly transforming interactions with multimedia content, enabling intuitive natural-language prompting for diverse perception tasks~\cite{radford2021clip,kirillov2023sam,ravi2024sam2}. This shift is particularly impactful for content-based retrieval and video understanding, where bridging the semantic gap between pixels and language is paramount.
SAM3~\cite{carion2025sam3segmentconcepts} extends this paradigm to concept-driven segmentation, allowing users to query and isolate specific regions using free-form textual prompts.
While pivotal for fine-grained analysis of complex scenes, SAM3 inherits a large CLIP-style text encoder originally designed for broad vision-language understanding. This introduces a fundamental architectural mismatch: segmentation prompts are typically short noun phrases describing visual concepts, e.g., \texttt{white dog} or \texttt{person in a blue jacket}, 
while the encoder employs deep transformer stacks and imposes a significant persistent memory overhead ($>$350MB) to support a level of syntactic complexity that segmentation prompts rarely require. Although text encoding latency is often amortized in video processing, its static model VRAM footprint remains a prohibitive bottleneck for edge hardware, where every megabyte counts.
On the other hand, the prevailing trend in efficient vision models has largely focused on optimizing the image encoder~\cite{xiong2024efficientsam,zhang2023mobilesam}. Yet, in vision-language tasks, the text encoder is often treated as a fixed component. We argue that this asymmetric optimization leaves significant efficiency gains on the table, especially for devices with strict memory budgets.

With this paper, we seek to answer the following question: \emph{To what degree can SAM 3's text encoder capacity be reduced without compromising grounding accuracy, specifically within domains characterized by high prompt redundancy?}

Based on a large-scale anatomical analysis of 404,796 prompts, we identify that vision-language segmentation prompts fundamentally differ from general web text: they operate on a low-dimensional manifold (intrinsic dimensionality $\approx 16$) and rely on sparse object-centric vocabulary.
Leveraging this domain-specific redundancy, we propose \textbf{SAM3-LiteText}, a streamlined text encoding framework that replaces the heavy SAM3 encoder with lightweight MobileCLIP students.
Unlike generic knowledge distillation~\cite{hinton2015distilling}, our method explicitly incorporates domain constraints: we optimize the context window to the prompt distribution ($L=16$) and introduce a \textit{consistency loss} that enforces permutation invariance, treating prompts as ``bags of concepts'' robust to syntactic variations. The primary contributions of this work are summarized below:

\begin{itemize}[leftmargin=*]
    \item \textbf{Anatomical Prior for Compression:} we provide the first quantitative characterization of text encoder redundancy in segmentation, showing that 75.5\% of context and 90\% of embedding capacity are superfluous.
    \item \textbf{SAM3-LiteText Framework:} we propose a domain-aware distillation strategy that enforces permutation invariance and collapses the context window, enabling aggressive compression without performance collapse.
    \item \textbf{Efficient Deployment:} we demonstrate that our 42M parameter student (with a $88\%$ reduction in model size) retains $98.1\%$ of the teacher's performance, effectively solving the static memory bottleneck for on-device segmentation.
\end{itemize}

By significantly reducing the computational and memory footprint of vision-language models, SAM3-LiteText represents a meaningful step forward in democratizing access to advanced AI capabilities on resource-constrained edge devices. This efficiency not only enables privacy-preserving applications in robotics and mobile computing but also contributes to a lower carbon footprint for large-scale AI deployment. We hope this work inspires further research into domain-aware model compression, leading to more sustainable and accessible foundation models for the broader community.
\section{Related Work}
\label{sec:related}

\paragraph{Vision-Language Segmentation. }
The integration of natural language with image segmentation has progressed from referring expression comprehension~\cite{hu2016segmentation} to open-vocabulary segmentation~\cite{ghiasi2022scaling,xu2023open} and universal segmentation models~\cite{kirillov2023sam,ravi2024sam2}.
SAM~\cite{kirillov2023sam} introduced the promptable segmentation paradigm, later extended by SAM2~\cite{ravi2024sam2} and SAM3~\cite{carion2025sam3segmentconcepts} with text-based prompting employing CLIP-style~\cite{radford2021clip} text encoders.
The DETR paradigm~\cite{carion2020end} evolved to support text conditioning via MDETR~\cite{kamath2021mdetr}, enabling end-to-end multi-modal understanding. More recently, Grounding DINO~\cite{liu2023grounding} combines the DINO detector with grounded pre-training for open-set object detection, while GLEE~\cite{wu2024glee} unifies multiple prompting modalities for general object foundation modeling. Recent work has also integrated SAM with CLIP for enhanced semantic understanding~\cite{wang2024samclip}.
Although these models achieve impressive zero-shot performance, their computational complexity remains prohibitive for edge deployment.

\paragraph{Multi-Object Tracking (MOT) and Segmentation. }
Traditional MOT methods employ tracking-by-detection, associating independent frame-level detections using motion and appearance cues~\cite{bewley2016simple,wojke2017simple,zhang2022bytetrack}.
Alternatively, end-to-end transformer architectures jointly optimize detection and association~\cite{meinhardt2022trackformer,zeng2022motr,yu2023motrv3}.
Recently, SAM2MOT~\cite{jiang2025sam2mot} and SAMURAI~\cite{yang2024samurai} have adapted SAM2 for zero-shot tracking based on memory mechanisms.
Video object segmentation benchmarks like DAVIS~\cite{pont20172017} and YouTube-VOS~\cite{xu2018youtubevos} have driven progress in this area.
SAM3 extends these capabilities to concept-driven segmentation, highlighting the importance of efficient text encoding for real-time video applications.

\paragraph{Efficient Vision Foundation Models. }
Significant effort has been devoted to compressing the image encoder of SAM.
EfficientSAM~\cite{xiong2024efficientsam} leverages masked image pretraining for efficient variants, while FastSAM~\cite{zhao2023fast} reformulates segmentation as a CNN-based detection task. Mobile-SAM~\cite{zhang2023mobilesam} and EdgeSAM~\cite{zhou2024edgesam} target mobile deployment through architectural innovations and prompt-in-the-loop distillation.
RepViT-SAM~\cite{wang2024repvit} achieves real-time performance by employing an efficient backbone design.
For SAM2 video segmentation, multi-teacher distillation approaches~\cite{zeng2025multiteacher} have shown promising results in balancing efficiency and accuracy.
Most recently, EfficientSAM3~\cite{zeng2025efficientsam3} introduced progressive hierarchical distillation across SAM generations, achieving substantial speedups for video concept segmentation.
Domain-specific adaptations have also emerged, including efficient SAM variants for medical imaging applications~\cite{zeng2025agglomerating, ma2024medsam} and test-time prompt-guided training for specialized segmentation tasks~\cite{zeng2025tuning}.
However, these works focus predominantly on the \emph{image encoder}, leaving the text encoder, which accounts for significant latency in text-prompted scenarios, largely unexplored.

\paragraph{Efficient Text Encoders. }
Prior work on efficient text encoding focuses on general NLP tasks.
DistilBERT~\cite{sanh2019distilbert} and TinyBERT~\cite{jiao2020tinybert} apply knowledge distillation to compress BERT, while MobileBERT~\cite{sun2020mobilebert} designs mobile-friendly architectures.
MobileCLIP~\cite{vasu2024mobileclip} specifically targets vision-language applications, achieving significant speedups through hybrid CNN-Transformer designs.
However, these approaches treat text encoding as a general-purpose task without exploiting the domain-specific properties of segmentation prompts.
Our work complements existing image encoder compression efforts by providing the first systematic study of text encoder redundancy in vision-language segmentation.

\section{Anatomical Analysis: Quantifying Text Encoder Redundancy}
\label{sec:analysis}

To quantify redundancy within the text encoder and identify the allocation of computation, we first conduct a systematic anatomical study of the SAM3 text encoder path
\texttt{(tokenizer $\to$ embedder $\to$ text encoder)}. In this study, 404,796 unique text prompts from six diverse sources are analyzed to answer three fundamental questions: (1) how much of the context window is actually used? (2) how much of the embedding space is exploited? (3) what is the true dimensionality of the output manifold?
The results of our analysis reveal severe over-provisioning across all three dimensions.

\subsection{Prompt Statistics and Context Window}
\label{sec:prompt_stats}

We first characterize the statistical properties of the noun prompts, which fundamentally differ from general NLP corpora.

\subsubsection{Datasets and Preprocessing}

We aggregate prompts from six diverse segmentation benchmarks that represent different annotation paradigms: \textit{RF100-VL}~\cite{rf100} (554 category names), \textit{LVIS}~\cite{gupta2019lvis} (1,199 category names), \textit{RefCOCO}~\cite{yu2016refcoco,mao2016refcocog} (246,040 referring expressions), and the \textit{SA-Co} suite~\cite{carion2025sam3segmentconcepts} comprising Gold (51,577 human-annotated descriptions), Silver (54,178 semi-automatic annotations), and VEval (51,248 evaluation annotations).

After deduplication, we obtain 404,796 unique prompts covering the entire spectrum from simple category names (\texttt{person}, mean token count $\mu$=3, including start token \texttt{SOT} and end token \texttt{EOT}) to descriptive phrases (\texttt{large white dog in the garden}, $\mu$=5.6 tokens) to complex referring expressions (\texttt{the man in the red shirt on the left}, $\mu$=9.4 tokens).
All prompts are tokenized using the SAM3 BPE tokenizer with a vocabulary size of 49,408 and a context length of $L=32$.

\subsubsection{Token Length Distribution}

\begin{figure}[t]
    \centering
    \includegraphics[width=\columnwidth]{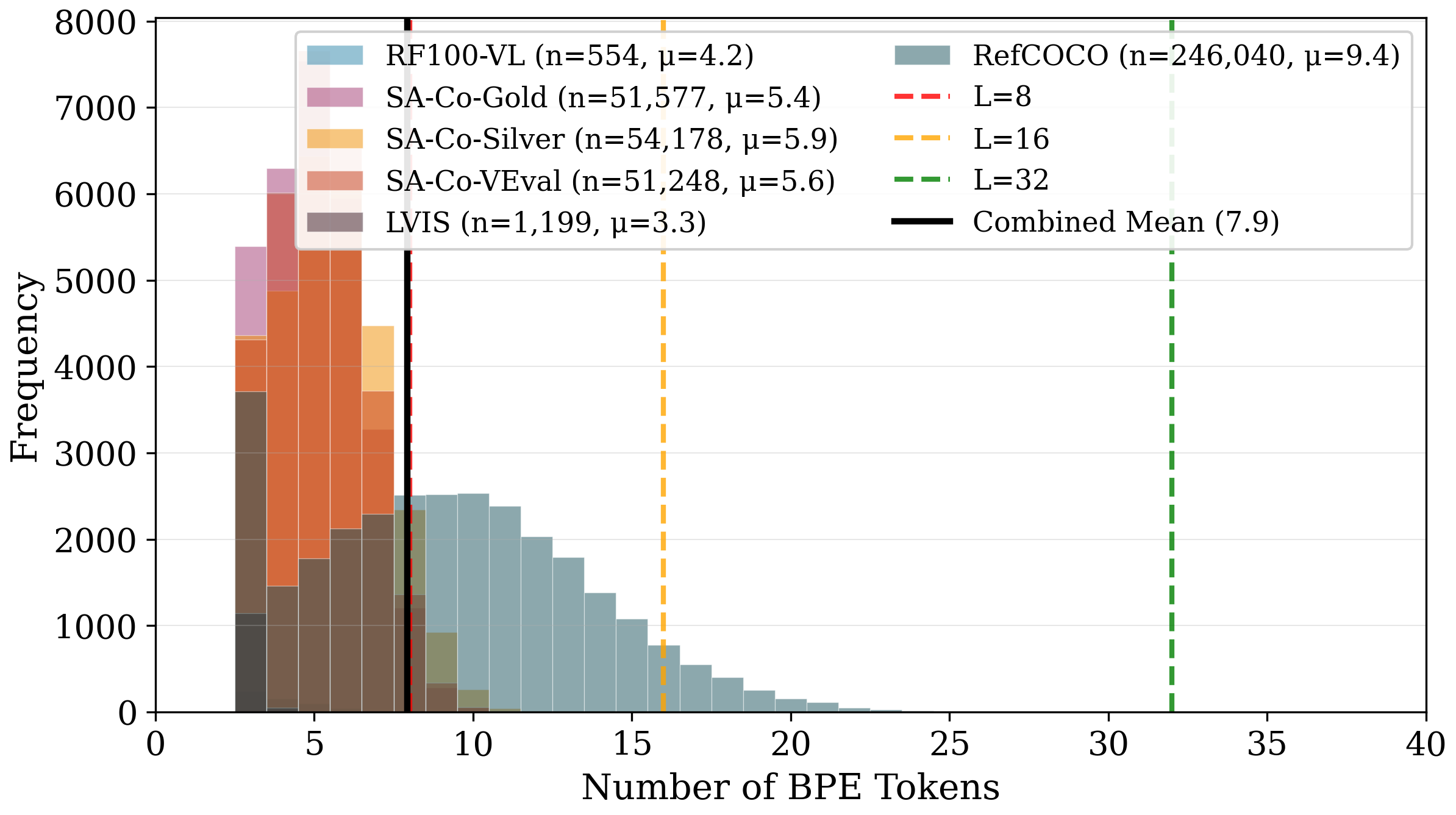}
    \caption{Token length distribution across 404,796 unique prompts from six sources: RF100-VL, SA-Co-Gold, SA-Co-Silver, SA-Co-VEval, LVIS, and RefCOCO. Each dataset component is shown separately, revealing distinct prompt length characteristics. The combined mean is $\mu$=7.9 tokens.}
    \label{fig:token_distribution}
    \Description{Overlaid histograms showing token length distributions for all six dataset components.}
\end{figure}

Figure~\ref{fig:token_distribution} visualizes the token length distribution across all six datasets, which reveals distinct prompt characteristics: LVIS category names are the shortest (mean token count $\mu$=3.3), SA-Co datasets are moderate ($\mu$=5.4--5.9), and RefCOCO referring expressions are the longest ($\mu$=9.4).
The combined mean of prompt length is $\mu = 7.9$ tokens, confirming that segmentation prompts are fundamentally different from general NLP text: they are short- to medium-length noun phrases rather than full sentences.
This observation has profound implications for encoder design - complex syntactic modeling may be unnecessary.

\subsubsection{Context Window Efficiency}

To quantify redundant computation, we define \emph{information density} as the ratio between content tokens and total context length: $\text{InfoDensity}(L) = \frac{1}{N}\sum_{i=1}^{N} \frac{\min(|t_i|, L)}{L}$, where $|t_i|$ is the token count for prompt $i$, $L$ is the context length, and $N$ is the total number of prompts.
Higher values indicate more efficient context utilization.
Additionally, \emph{truncation rate} is a prompt-level metric counting prompts are cut off across all datasets, while \emph{token loss} is a token-level metric measuring discarded tokens.

\begin{table}[t]
\caption{Context-window utilization across lengths $L$ on the full prompt audit (404,796 unique prompts). $L=32$ wastes most positions on padding, while $L=16$ substantially improves utilization with limited truncation, motivating $L=16$ for distillation.}
\label{tab:context_window}
\centering
\begin{tabular}{lcccr}
\toprule
Context $L$ & Info Density & Padding & Truncation & Token Loss \\
\midrule
32 & 0.245 & 75.5\% & 0.1\% & --- \\
16 & 0.480 & 52.0\% & 5.0\% & 2.1\% \\
8 & 0.800 & 20.0\% & 28.5\% & 12.3\% \\
\bottomrule
\end{tabular}
\end{table}
Table~\ref{tab:context_window} reveals the extent of computational redundancy in the combined dataset.
At the default $L$=32, information density is 0.245, indicating that 75.5\% of all attention computation is allocated to padding tokens.
Given the $O(L^2)$ complexity of self-attention mechanisms, this inefficiency scales quadratically with sequence length.
The truncation behavior varies by dataset component: at $L$=16, LVIS, RF100-VL, and SA-Co prompts experience near-zero truncation ($<$0.1\%), while RefCOCO referring expressions sees a 8.1\% truncation due to their longer average length.
This indicates that a context length of $L$=16 provides an optimal trade-off, with negligible truncation for short prompts and acceptable loss for longer expressions.

\subsubsection{Vocabulary Coverage}

Beyond sequence length, we analyze vocabulary utilization.
Of the 49,408 BPE tokens, only 17,300 (35.0\%) appear in our corpus, indicating that 65\% of the specific concepts available in the tokenizer are never referenced in segmentation prompts.

\begin{figure}[t]
    \centering
    \includegraphics[width=\columnwidth]{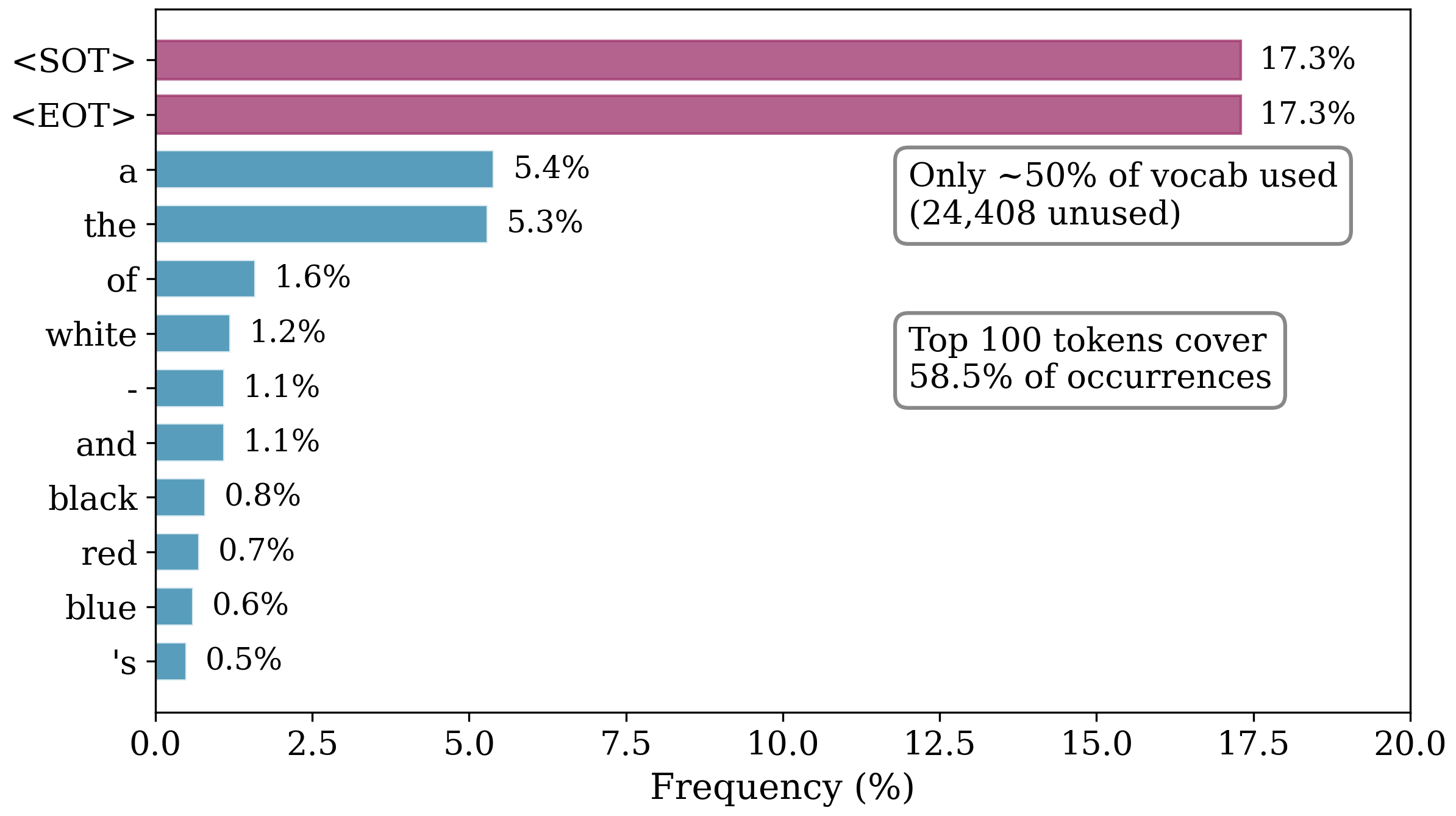}
    \caption{Vocabulary coverage analysis. (a) Only 35\% of the 49,408 BPE tokens are ever used in segmentation prompts. (b) Token frequency is highly skewed---the top 100 tokens cover 58.5\% of all occurrences.}
    \label{fig:vocab_coverage}
    \Description{Pie chart and bar chart showing vocabulary usage patterns.}
\end{figure}

As shown in Figure~\ref{fig:vocab_coverage}, the token distribution exhibits extreme skew: special tokens (\texttt{SOT}, \texttt{EOT}) account for 34.6\% of occurrences, while the top 100 most frequent tokens cover 58.5\% of the total volume.
The dominant vocabulary primarily consists of functional determiners (\texttt{a}, \texttt{the}, \texttt{of}) alongside visual attributes such as colors (\texttt{white}, \texttt{black}, \texttt{red}, \texttt{blue}) and size modifiers (\texttt{large}, \texttt{small}), further emphasizing the structured, repetitive nature of these prompts compared to open-domain text. This resonates with the work in video event detection, where exploiting concept distribution through the data shuffle~\cite{mettes2016imagenet} significantly improved representation learning.

\begin{tcolorbox}[findingbox]
\textit{\textbf{Finding~\RNum{1}.}} Segmentation prompts are short ($\mu$=7.9 tokens across 404K samples) and use sparse vocabulary (35\%), resulting in 75\% padding waste at $L$=32---the text encoder processes mostly empty tokens even with complex referring prompts.
\end{tcolorbox}

\subsection{Embedding Space Analysis}
\label{sec:embedding_analysis}

Having established that prompts are short and the vocabulary is sparse, we now examine whether the embedding representations themselves contain redundancy.
We analyze two components: (1) the token embedding matrix and (2) the positional embedding matrix.

\subsubsection{Token Embedding SVD Analysis}

The token embedding is a lookup table $E \in \mathbb{R}^{49408 \times 1024}$ mapping each vocabulary token to a 1024-dimensional vector.
If many singular values are small, the embedding exists on a lower-dimensional manifold and could be factorized as:
\begin{equation}
    E_{[V \times d]} \approx U_{[V \times k]} \cdot S_{[k \times k]} \cdot W_{[k \times d]}^\top \quad (k \ll d)
\end{equation}

We perform the singular value decomposition (SVD) for the SAM3 teacher's token embedding (centered by subtracting the mean) and analyze both the full vocabulary and the subset of 17,300 used tokens.

\begin{table}[t]
\caption{Token embedding rank utilization (SVD). Although the prompt vocabulary is sparse, the learned token embeddings remain high-rank, suggesting that pruning unused tokens is more promising than low-rank factorization.}
\label{tab:embedding_svd}
\centering
\begin{tabular}{lrrrr}
\toprule
Vocabulary & Eff. Rank & Rank \% & 90\% Var & 95\% Var \\
\midrule
Full (49.4K) & 943 & 92.1\% & 857 dims & 933 dims \\
Used (17.3K) & 906 & 88.5\% & 834 dims & 919 dims \\
\bottomrule
\end{tabular}
\end{table}

Table~\ref{tab:embedding_svd} indicates that the embedding dimension is well-utilized.
Even when restricting the analysis to the subset of vocabulary used, the effective rank remains high (906/1,024, or 88.5\%), and capturing 90\% of the variance requires 834 dimensions.
This suggests that the token embeddings in SAM3 exhibit a high degree of orthogonality, likely resulting from the CLIP-style contrastive pretraining, which pushes distinct concepts apart in the embedding space. Similar observations regarding semantic concepts have been made in multimedia literature, where this high-dimensional structure was leveraged for effective compression to facilitate interactive retrieval on a large scale of images~\cite{8048559}.
Consequently, unlike the positional embeddings discussed next, the token embedding matrix is information-dense and not amenable to significant low-rank compression without loss of semantic fidelity.

\subsubsection{Positional Embedding Similarity Analysis}

The text encoder uses a learnable positional embedding matrix $P \in \mathbb{R}^{32 \times 1,024}$. The $i$-th row of this matrix contains a dense vector representation for the sequence index $i$, which is added to the token embedding:
\begin{equation}
    \mathbf{x}_i = E[\text{token}_i] + P[i]
\end{equation}

Since most prompts use only positions 0--7 (mean $\sim$5.6 tokens), positions 8--31 receive \emph{far fewer gradient updates} during training.
We hypothesize that these under-trained positions have converged to similar values, rendering them redundant.

\begin{figure}[t]
    \centering
    \includegraphics[width=0.65\columnwidth]{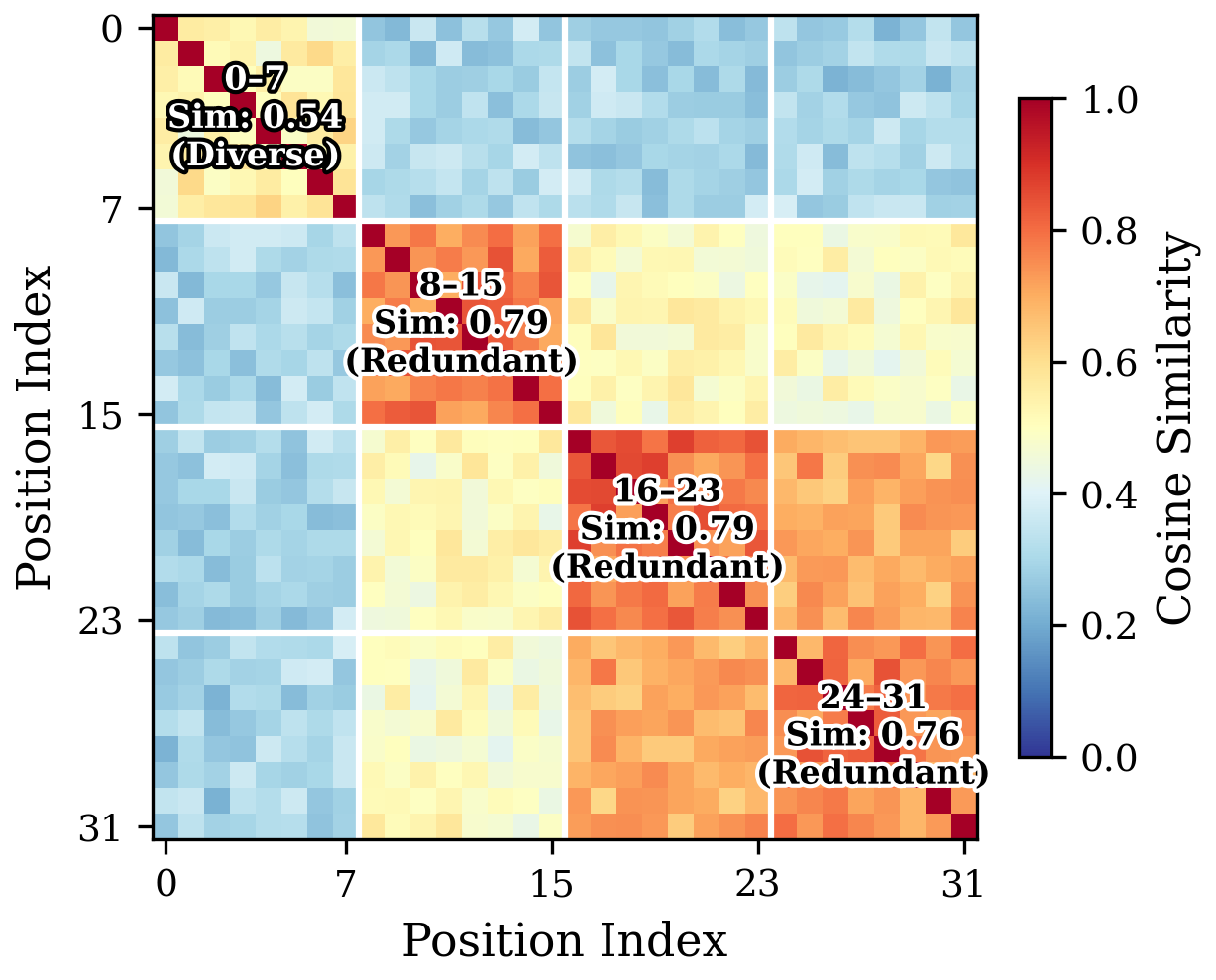}
    \caption{Positional embedding similarity analysis. The cosine similarity heatmap shows high correlation among late positions (8+), which are 1.5$\times$ more similar within-group than positions 0--7.}
    \label{fig:positional_similarity}
    \Description{Heatmap and bar chart showing positional embedding redundancy.}
\end{figure}

Figure~\ref{fig:positional_similarity} confirms this hypothesis.
We compute pairwise cosine similarity between all positional embedding vectors.
Positions 0--7 exhibit moderate within-group similarity (0.54), indicating diversity from training on real data.
In contrast, positions 8+ are $\sim$1.5$\times$ more similar to each other than positions 0--7 (0.76--0.79), indicating training-induced redundancy that could collapse without information loss.

SVD analysis of the 32$\times$1024 positional matrix reveals that 90\% of the variance is captured by only 7 principal components, with an effective rank of 20/32.
This strongly validates the truncation of the context window to $L$=8 or $L$=16.

\begin{tcolorbox}[findingbox]
\textit{\textbf{Finding~\RNum{2}.}} SAM3 Token embeddings are well-utilized (88\% effective rank), but positional embeddings for positions 8--31 are highly redundant (1.5$\times$ higher similarity).
\end{tcolorbox}

\subsection{Text Encoder Intrinsic Dimensionality}
\label{sec:intrinsic}

The final component of our anatomical study examines the output features of the text encoder.
The SAM3 text encoder produces 256-dimensional embeddings (after the resizer projection) that are used for cross-attention with image features.
To determine the true number of degrees of freedom in the learned representation, we analyze the \emph{intrinsic dimensionality} (ID) using two established estimators: Two-Nearest-Neighbors (TwoNN)~\cite{facco2017twonn}, which is based on ratios of distances to the two nearest neighbors and is robust to global structure, and Maximum Likelihood Estimation (MLE)~\cite{levina2004mle}, which estimates local dimensionality using $k$-nearest neighbor statistics (averaged over $k \in \{5, 10, 20, 50\}$). We extract output embeddings for 5,000 randomly sampled prompts from the SAM3 text encoder and compute the ID estimates.

\begin{figure}[t]
    \centering
    \includegraphics[width=1.0\columnwidth]{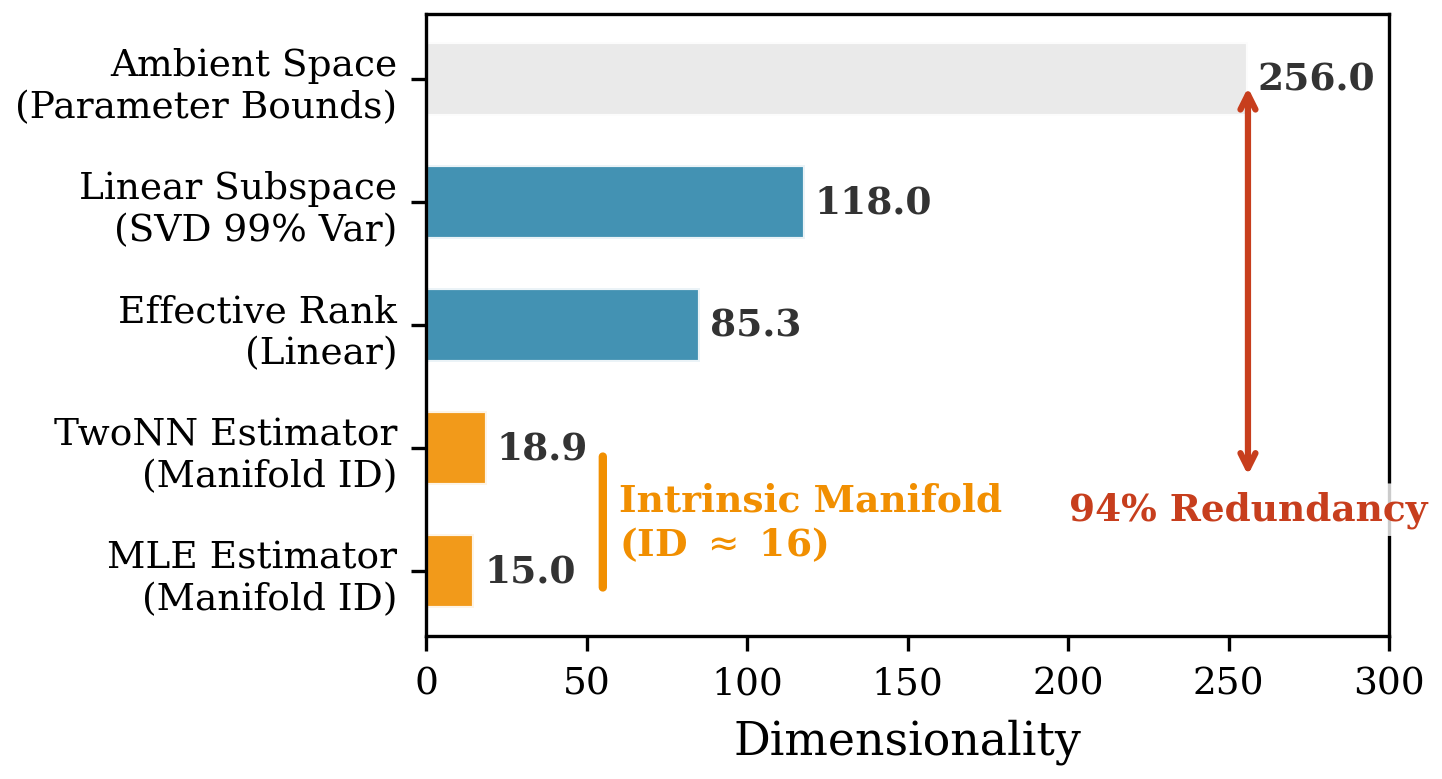}
    \caption{Output embedding intrinsic dimensionality. Utilization estimates indicate that only $\sim$6--8\% of the 256-dimensional space is actually used.}
    \label{fig:intrinsic_dim}
    \Description{Bar charts showing intrinsic dimensionality comparison and output space utilization.}
\end{figure}

Figure~\ref{fig:intrinsic_dim} quantitatively illustrates the hierarchy of dimensionality within the SAM3 Teacher's output embeddings, providing a geometric explanation for the potential compressibility. We observe a striking ``collapse'' in dimensionality as we transition from the theoretical parameter bounds to the actual data geometry.

While the Ambient Space is fixed at $d=256$ by design, linear analysis based on Singular Value Decomposition (SVD) reveals an Effective Rank of only 85.3. This linear metric suggests that the data ellipsoid is already significantly flattened, utilizing only a third of the available linear subspaces. However, non-linear estimators reveal an even tighter structure. Both TwoNN and MLE estimators converge on an Intrinsic Dimensionality (ID) of approximately 16 ($\pm 2$). The strong agreement between these two distinct estimators---one based on neighbor distance ratios and the other on likelihood maximization---reinforces the validity of this finding.

This massive discrepancy---where $\text{ID} \ll d_{\text{eff}} < d$---indicates that approximately 94\% of the output space is redundant capacity. The embeddings do not fill the 256-dimensional volume but rather lie on a thin, low-dimensional non-linear manifold embedded within it. This supports the Manifold Hypothesis in the context of specific domain prompts: while the space of all possible English sentences might be high-dimensional, the subspace of ``segmentation prompts'' (noun phrases describing physical objects) is surprisingly compact. This finding is pivotal for our distillation strategy: it confirms that a student architecture with a significantly smaller output dimension (e.g., $d=32$) effectively possesses sufficient topological capacity to capture the teacher's true semantic signal without much information loss.

\begin{tcolorbox}[findingbox]
\textit{\textbf{Finding~\RNum{3}.}} Despite the 256-dimensional ambient space, the SAM3 text encoder's features reside on a low-dimensional manifold with an intrinsic dimensionality of $\sim$16--19.
\end{tcolorbox}

Our anatomical analysis reveals that the SAM3 text encoder is massively over-provisioned for vision-language segmentation.

\section{Method: Knowledge Distillation for Efficient Text Encoding}
\label{sec:method}
Our anatomical analysis in Section~\ref{sec:analysis} exposes a significant mismatch between the capacity of the SAM 3 text encoder and the actual requirements of the segmentation task. Motivated by this observation, we propose SAM3-LiteText, the first distilled approach for SAM 3 that explicitly exploits spectral and attentional redundancy. SAM3-LiteText is built upon three novel technical contributions derived directly from our analysis:

\begin{enumerate}[leftmargin=*]
\item \textbf{Spectral Compression:} We map the student architecture to the teacher's intrinsic dimensionality. Since Finding \RNum{3} shows that the output manifold collapses to a low dimension, we demonstrate that efficient MobileCLIP variants effectively saturate the required information bound.
\item \textbf{Spatiotemporal Attention Cutting:} We define a tailored context window of $L=16$ based on the coverage analysis in Finding~\RNum{1}. This architectural constraint hard-codes the observation that tokens beyond position 8 yield negligible gain (Finding~\RNum{2}), optimizing the attention mechanism at the structural level.

\item \textbf{Semantic Regularization:} We propose a ``Bag of Concepts'' prior. Unlike standard distillation that mimics the teacher's exact logits, our method utilizes a consistency loss that rewards permutation invariance, filtering out the syntactic noise in the teacher's language modeling objective.
\end{enumerate}

\subsection{Student Architecture}

We distill from the Teacher model SAM3 text encoder (CLIP ViT-L/14, 353.72M parameters) to three MobileCLIP variants:
\begin{itemize}[leftmargin=*]
    \item \textbf{MobileCLIP-S0}: 4 transformer layers, 512-dim embeddings, 42.55M parameters (88\% reduction)
    \item \textbf{MobileCLIP-S1}: 12 transformer layers, 512-dim embeddings, 63.54M parameters (82\% reduction)
    \item \textbf{MobileCLIP2-L}: 12 transformer layers, 768-dim embeddings, 123.81M parameters (65\% reduction)
\end{itemize}

All three student encoders use the same tokenizer as the teacher (49,408 BPE vocabulary) and output 256-dimensional embeddings (via a projection layer) to maintain compatibility with the frozen SAM3 fusion modules.
This architecture directly exploits Finding~\RNum{3}: since the true signal lies on a low-dimensional manifold, the smaller widths (512/768 vs 768) and shallower depths (4/12 vs 12) of MobileCLIP are sufficient to capture the necessary semantics. 

With a context length $L=16$, the parameter counts are 42.54M (S0), 63.53M (S1), and 123.80M (2-L). This 88\% parameter reduction (for S0) is critical for edge deployment, as the static VRAM footprint is often a harder constraint than latency for non-streaming prompts.

\subsection{Context Length Optimization}

We train and evaluate student models with a reduced context length of $L=16$.
This decision is strictly guided by Finding~\RNum{1}, which showed that $L=32$ results in 75.5\% padding waste, and Finding~\RNum{2}, which revealed a high redundancy in positional embeddings beyond index 8.
By setting $L=16$, we reduce the computational overhead of processing padding tokens while incurring only 2.1\% token truncation loss.
Crucially, we train the student \emph{from scratch} with this reduced window, preventing it from learning to attend to empty positions.

\subsection{Training Objective}

We train the student encoders through knowledge distillation from the frozen SAM3 teacher using the following loss:
\begin{equation}
    \mathcal{L}_{\text{total}} = \mathcal{L}_{\text{MSE}} + \lambda_1 \mathcal{L}_{\text{cos}} + \lambda_2 \mathcal{L}_{\text{consist}},
\end{equation}
where the \textit{MSE loss} $\mathcal{L}_{\text{MSE}}$ aims to achieve coordinate alignment for compatibility with pretrained fusion modules:
\begin{equation}
    \mathcal{L}_{\text{MSE}} = \|\mathbf{v}_s - \mathbf{v}_t\|^2,
\end{equation}
in which $\mathbf{v}_s$ and $\mathbf{v}_t$ are the student and teacher embeddings.

The \textit{cosine loss} $\mathcal{L}_{\text{cos}}$ is designed to perform direction alignment for semantic similarity:
\begin{equation}
    \mathcal{L}_{\text{cos}} = 1 - \frac{\mathbf{v}_s \cdot \mathbf{v}_t}{\|\mathbf{v}_s\| \|\mathbf{v}_t\|}.
\end{equation}

The \textit{consistency loss} $\mathcal{L}_{\text{consist}}$ is employed to enforce the bag-of-concepts property and improve robustness to word ordering. We minimize the $L_2$ distance between the student embedding of the original prompt $\mathbf{v}_s(T)$ and the embedding of a syntactically permuted version $\mathbf{v}_s(T')$:
\begin{equation}
    \mathcal{L}_{\text{consist}} = \|\mathbf{v}_s(T) - \mathbf{v}_s(T')\|^2.
\end{equation}
Here, $T'$ denotes the modified prompt where attribute-noun order is shuffled (e.g., \texttt{white shirt man} $\rightarrow$ \texttt{man white shirt}). This regularization forces the student model to learn representations that are invariant to such trivial syntactic permutations.

\section{Experiments}
\label{sec:experiments}

We evaluate whether the proposed anatomical findings translate into practical efficiency gains without degrading downstream grounding and segmentation.
In addition to reporting end-task metrics, we measure distillation quality (embedding alignment to the teacher) and end-to-end efficiency (latency/FPS of the text encoder under the same batch and context settings).

\subsection{Experimental Setup}

\paragraph{Datasets.}
We performed the student training on a combined corpus of prompts from four datasets: RF100-VL~\cite{rf100} (554 unique category names) and RefCOCO/RefCOCO+/RefCOCOg~\cite{yu2016refcoco,mao2016refcocog} (246,040 referring expressions).
This training set covers a spectrum from simple category names to complex referring expressions, providing diverse prompt variations for knowledge distillation.
While our anatomical analysis (Section~\ref{sec:analysis}) examines 404,796 prompts from six sources, including SA-Co Gold/Silver/VEval and LVIS datasets, we only use RefCOCO/RefCOCO+/RefCOCOg and RF100-VL for training the distilled text encoders for a fair benchmark.
For downstream segmentation evaluation, we use SA-Co Gold~\cite{carion2025sam3segmentconcepts} for instance segmentation and SA-Co VEval~\cite{carion2025sam3segmentconcepts} for video object segmentation.

\paragraph{Baselines.}
We benchmark the distilled student models against the following baselines:
 the original SAM3~\cite{carion2025sam3segmentconcepts}, gDino-T~\cite{liu2023grounding}, OWLv2 \cite{minderer2024scalingopenvocabularyobjectdetection}, LLMDet-L~\cite{fu2025llmdet}, APE-D~\cite{Shen_2024_CVPR}, DINO-X~\cite{ren2025dinoxunifiedvisionmodel}, and Gemini 2.5~\cite{google2025gemini25}.

\paragraph{Metrics.}
We report CG\_F1 (classification-gated F1) as our primary metric, computed as $\text{CG\_F1} = 100 \cdot \text{pmF}_1 \cdot \text{IL\_MCC}$.
Here, pmF1 (positive micro F1) measures localization quality on positive pairs, while IL\_MCC (Image-Level Matthews Correlation Coefficient) evaluates binary presence classification~\cite{carion2025sam3segmentconcepts}.
For video tasks, we use VL\_MCC (Video-level MCC) to compute CG\_F1, along with pHOTA (phrase-based HOTA), pDetA (detection accuracy), pAssA (association accuracy), and TETA (Track Every Thing Accuracy).
We also report model efficiency (parameters, FLOPs, latency).

\paragraph{Implementation Details.}
Student models were trained for 100 epochs using the AdamW optimizer~\cite{loshchilov2017decoupled} with a base learning rate of $5 \times 10^{-3}$, a weight decay of 0.05, and a cosine learning rate scheduler~\cite{loshchilov2016sgdr} with 10 warmup epochs. We used a batch size of 64 with 2 gradient accumulation steps. The loss function combined MSE loss ($\lambda_{\text{MSE}}=1.0$) and cosine similarity loss ($\lambda_{\text{cos}}=2.0$). For models using consistency regularization, we set $\lambda_{\text{consist}}=0.1$. Training and evaluation were performed on 4 $\times$ NVIDIA H200 GPUs with mixed precision (AMP) enabled, while inference speed benchmarks were conducted on a single NVIDIA RTX 4070 Ti.

\subsection{Main Results}

\begin{table*}[t]
\caption{SA-Co Gold results comparing the SAM3 teacher and MobileCLIP student text encoders ($L=16$).}
\label{tab:main_results}
\centering
\tiny
\renewcommand{\arraystretch}{1.0}
\setlength{\tabcolsep}{3pt}
\resizebox{0.99\linewidth}{!}{\begin{tabular}{r|rrr|rrr|rrr|rrr|rrr|rrr|rrr|rrr}
\toprule
\multirow{2}{*}{\textbf{Model}} & \multicolumn{3}{c|}{\textbf{Avg}} & \multicolumn{3}{c|}{\textbf{MC}} & \multicolumn{3}{c|}{\textbf{SA1B}} & \multicolumn{3}{c|}{\textbf{Crd}} & \multicolumn{3}{c|}{\textbf{F\&D}} & \multicolumn{3}{c|}{\textbf{Spt}} & \multicolumn{3}{c|}{\textbf{Att}} & \multicolumn{3}{c}{\textbf{Wiki}} \\
\cmidrule{2-25}
& \textbf{C} & \textbf{M} & \textbf{P} & \textbf{C} & \textbf{M} & \textbf{P} & \textbf{C} & \textbf{M} & \textbf{P} & \textbf{C} & \textbf{M} & \textbf{P} & \textbf{C} & \textbf{M} & \textbf{P} & \textbf{C} & \textbf{M} & \textbf{P} & \textbf{C} & \textbf{M} & \textbf{P} & \textbf{C} & \textbf{M} & \textbf{P} \\
\midrule
gDino-T & 3.3 & 0.15 & 16.2 & 2.9 & 0.21 & 13.9 & 3.1 & 0.20 & 15.4 & 0.28 & 0.08 & 3.4 & 0.96 & 0.10 & 9.8 & 1.1 & 0.10 & 11.2 & 13.8 & 0.29 & 47.3 & 0.70 & 0.06 & 12.1 \\
OWLv2 & 17.3 & 0.46 & 36.8 & 12.2 & 0.39 & 31.3 & 9.8 & 0.45 & 21.7 & 8.9 & 0.36 & 24.8 & 24.4 & 0.51 & 47.9 & 24.4 & 0.52 & 47.0 & 25.9 & 0.54 & 48.2 & 15.4 & 0.42 & 36.6 \\
LLMDet-L & 6.5 & 0.21 & 27.3 & 4.5 & 0.23 & 19.4 & 5.3 & 0.23 & 22.8 & 2.4 & 0.18 & 13.7 & 5.5 & 0.19 & 29.1 & 4.4 & 0.17 & 25.3 & 22.2 & 0.39 & 57.1 & 1.2 & 0.05 & 23.3 \\
APE-D & 16.4 & 0.40 & 36.9 & 12.6 & 0.42 & 30.1 & 2.2 & 0.22 & 10.0 & 7.2 & 0.35 & 20.3 & 22.7 & 0.51 & 45.0 & 31.8 & 0.56 & 56.5 & 26.7 & 0.47 & 57.3 & 11.6 & 0.29 & 39.5 \\
DINO-X & 21.3 & 0.38 & 55.2 & 17.2 & 0.35 & 49.2 & 19.7 & 0.48 & 40.9 & 12.9 & 0.34 & 37.5 & 30.1 & 0.49 & 61.7 & 28.4 & 0.41 & 69.4 & 31.0 & 0.42 & 74.0 & 9.7 & 0.18 & 53.5 \\
Gemini 2.5 & 13.0 & 0.29 & 46.1 & 9.9 & 0.29 & 33.8 & 13.1 & 0.41 & 32.1 & 8.2 & 0.27 & 30.3 & 19.6 & 0.33 & 59.5 & 15.1 & 0.28 & 53.5 & 18.8 & 0.30 & 63.1 & 6.5 & 0.13 & 50.3 \\
\midrule
SAM3 (Teacher) & 54.1 & 0.82 & 66.1 & 47.3 & 0.81 & 58.6 & 53.7 & 0.86 & 62.6 & 61.1 & 0.90 & 67.7 & 53.4 & 0.79 & 67.3 & 65.5 & 0.89 & 73.8 & 54.9 & 0.76 & 72.0 & 42.5 & 0.70 & 60.9 \\
\midrule
\rowcolor{rowblue} S3LT (MC-S0, $L$=16) & 51.9 & 0.79 & 63.6 & 45.4 & 0.78 & 56.3 & 51.6 & 0.82 & 60.2 & 58.8 & 0.86 & 65.2 & 51.3 & 0.76 & 64.8 & 63.0 & 0.85 & 71.1 & 52.7 & 0.73 & 69.3 & 40.7 & 0.67 & 58.6 \\
\rowcolor{rowblue} S3LT (MC-S1, $L$=16) & 52.5 & 0.80 & 64.3 & 45.9 & 0.79 & 56.9 & 52.1 & 0.83 & 60.8 & 59.4 & 0.87 & 65.8 & 51.8 & 0.77 & 65.5 & 63.7 & 0.86 & 71.8 & 53.3 & 0.74 & 70.1 & 41.2 & 0.68 & 59.2 \\
\rowcolor{rowblue} S3LT (MC2-L, $L$=16) & 53.1 & 0.80 & 64.9 & 46.3 & 0.79 & 57.5 & 52.7 & 0.84 & 61.5 & 60.0 & 0.88 & 66.5 & 52.4 & 0.77 & 66.1 & 64.3 & 0.87 & 72.6 & 53.9 & 0.74 & 70.8 & 41.6 & 0.68 & 59.8 \\
\bottomrule
\end{tabular}}
\end{table*}

We evaluate our distilled text encoders on the SA-Co Gold benchmark~\cite{carion2025sam3segmentconcepts}, which consists of 51,577 human-annotated text descriptions across seven diverse subsets: metaclip\_nps, sa1b\_nps, crowded, fg\_food, fg\_sports\_equipment, attributes, and wiki\_common.
We report three metrics: CG\_F1 (concept grounding F1 score), IL\_MCC (instance-level Matthews correlation coefficient), and pmF1 (per-mask F1 score).

Table~\ref{tab:main_results} presents comprehensive quantitative results on SA-Co Gold, comparing our SAM3-LiteText variants with baseline models including gDino-T, OWLv2, LLMDet-L, APE-D, DINO-X, and Gemini 2.5.
The SAM3 teacher with CLIP ViT-L/14 text encoder achieves state-of-the-art performance (CG\_F1: 54.1, IL\_MCC: 0.82, pmF1: 66.1), significantly outperforming all baseline models. It has demonstrated that all distilled MobileCLIP variants ($L=16$) maintain strong performance close to the teacher. The MobileCLIP-S0 variant achieves comparable results while reducing parameters by 88\% (42.54M vs 353.72M), confirming effective compression.

\subsection{Video Segmentation Results}
\begin{table*}[t]
\caption{Video segmentation and tracking on SA-V, YT-Temporal-1B, and SmartGlasses: replacing the SAM3 text encoder with distilled MobileCLIP students yields comparable performance to the teacher across metrics.}
\label{tab:video_results}
\centering
\tiny
\renewcommand{\arraystretch}{1.0}
\setlength{\tabcolsep}{3pt}
\resizebox{0.99\linewidth}{!}{%
\begin{tabular}{r|rrrrrrr|rrrrrrr|rrrrrrr}
\toprule
\multirow{2}{*}{\textbf{Model}} & \multicolumn{7}{c|}{\textbf{SA-V test (1.5K NPs)}} & \multicolumn{7}{c|}{\textbf{YT-Temporal-1B test (1.5K NPs)}} & \multicolumn{7}{c}{\textbf{SmartGlasses test (2.2K NPs)}} \\
\cmidrule{2-22}
& \textbf{C} & \textbf{M} & \textbf{P} & \textbf{H} & \textbf{D} & \textbf{A} & \textbf{T} & \textbf{C} & \textbf{M} & \textbf{P} & \textbf{H} & \textbf{D} & \textbf{A} & \textbf{T} & \textbf{C} & \textbf{M} & \textbf{P} & \textbf{H} & \textbf{D} & \textbf{A} & \textbf{T} \\
\midrule
Human & 53.1 & 0.87 & 60.8 & 70.5 & 55.5 & 90.2 & 69.1 & 71.2 & 0.97 & 73.2 & 78.4 & 68.8 & 89.7 & 84.3 & 58.5 & 0.92 & 63.5 & 72.3 & 58.8 & 89.4 & 77.1 \\
GLEE (all NPs) & 0.1 & 0.22 & 0.6 & 8.7 & 2.8 & 30.9 & 7.1 & 1.6 & 0.24 & 6.7 & 16.7 & 6.8 & 42.6 & 16.5 & 0.0 & 0.16 & 0.3 & 4.7 & 1.6 & 15.3 & 9.3 \\
LLMDet + SAM3 Tracker & 2.3 & 0.15 & 14.7 & 30.1 & 12.0 & 76.0 & 28.5 & 8.0 & 0.33 & 24.1 & 37.9 & 18.7 & 77.0 & 33.4 & 0.3 & 0.02 & 16.8 & 18.6 & 4.7 & 74.1 & 39.0 \\
SAM3 Detector + T-by-D & 25.7 & 0.72 & 35.8 & 55.7 & 40.2 & 78.3 & 53.2 & 47.6 & 0.93 & 51.3 & 68.2 & 60.8 & 77.0 & 70.4 & 29.7 & 0.85 & 35.1 & 60.0 & 50.6 & 71.7 & 61.2 \\
\midrule
SAM3 w/o temporal & 27.1 & 0.63 & 42.9 & 55.9 & 37.9 & 83.4 & 57.1 & 47.3 & 0.84 & 56.3 & 67.8 & 57.1 & 81.2 & 69.8 & 34.4 & 0.75 & 45.8 & 61.6 & 47.0 & 81.2 & 66.0 \\
SAM3 & \textbf{30.3} & 0.69 & \textbf{43.7} & \textbf{58.0} & \textbf{40.9} & \textbf{83.4} & 56.8 & \textbf{50.8} & 0.89 & \textbf{57.2} & \textbf{69.9} & 60.2 & \textbf{81.7} & \textbf{70.5} & \textbf{36.4} & 0.78 & \textbf{46.6} & \textbf{63.6} & 50.0 & \textbf{81.5} & 65.9 \\
\midrule
\rowcolor{rowblue} S3LT (MC-S0, $L$=16) & 29.4 & 0.67 & 42.4 & 56.2 & 39.7 & 81.6 & 55.1 & 49.3 & 0.86 & 55.5 & 67.8 & 58.4 & 79.2 & 68.4 & 35.3 & 0.76 & 45.2 & 61.7 & 48.5 & 79.1 & 63.9 \\
\rowcolor{rowblue} S3LT (MC-S1, $L$=16) & 29.7 & 0.68 & 42.8 & 56.8 & 40.1 & 82.2 & 55.7 & 49.8 & 0.87 & 56.1 & 68.5 & 59.0 & 80.1 & 69.1 & 35.7 & 0.77 & 45.7 & 62.3 & 49.0 & 79.9 & 64.6 \\
\rowcolor{rowblue} S3LT (MC2-L, $L$=16) & 30.0 & 0.68 & 43.3 & 57.4 & 40.5 & 82.8 & 56.2 & 50.3 & 0.88 & 56.6 & 69.2 & 59.6 & 80.9 & 69.8 & 36.0 & 0.77 & 46.1 & 63.0 & 49.5 & 80.7 & 65.2 \\
\bottomrule
\end{tabular}}
\end{table*}

Table~\ref{tab:video_results} presents video segmentation and tracking results on three test sets. It can be observed that
SAM3 achieves state-of-the-art performance on all three datasets, significantly outperforming baseline methods, including GLEE~\cite{wu2024glee} and LLMDet-based approaches~\cite{fu2025llmdet}.
Despite a lower model capacity, our lightweight models still maintain tracking metrics, which confirm the parameter redundancy within the original SAM3 teacher for these tasks.

\subsection{Ablation Studies}

We conduct comprehensive ablation studies to validate key design choices in our knowledge distillation approach.
All ablations are performed on the SA-Co Gold benchmark using MobileCLIP-S0 as the student architecture, unless otherwise noted.

\paragraph{Context Length Analysis.}
Table~\ref{tab:context_length_ablation} reports a context-length ablation on SA-Co Gold across all three MobileCLIP student architectures ($L\in\{8,16,32\}$).
The results show that matching the student to a shorter window is critical: using $L=32$ consistently degrades grounding across students, while $L=16$ recovers most of the teacher performance at a much lower attention cost.
Reducing further to $L=8$ introduces a small but consistent additional drop.

\begin{table}[t]
\caption{Context-length ablation on SA-Co Gold. Matching the student to a shorter window ($L=16$) has minimum performance degradation.}
\label{tab:context_length_ablation}
\centering
\footnotesize
\renewcommand{\arraystretch}{1.0}
\setlength{\tabcolsep}{3pt}
\begin{tabular}{c|rrr|rrr|rrr}
\toprule
\multirow{2}{*}{\textbf{Model}} & \multicolumn{3}{c|}{\textbf{$L=32$}} & \multicolumn{3}{c|}{\textbf{$L=16$}} & \multicolumn{3}{c}{\textbf{$L=8$}} \\
\cmidrule{2-10}
& \textbf{C} & \textbf{M} & \textbf{P} & \textbf{C} & \textbf{M} & \textbf{P} & \textbf{C} & \textbf{M} & \textbf{P} \\
\midrule
S3LT (MC-S0) & 52.2 & 0.79 & 63.9 & 51.9 & 0.79 & 63.6 & 50.3 & 0.77 & 61.7 \\
S3LT (MC-S1) & 52.8 & 0.80 & 64.6 & 52.5 & 0.80 & 64.3 & 50.9 & 0.78 & 62.4 \\
S3LT (MC2-L) & 53.4 & 0.80 & 65.2 & 53.1 & 0.80 & 64.9 & 51.5 & 0.78 & 63.0 \\
\bottomrule
\end{tabular}
\end{table}

The ablation results in Table~\ref{tab:context_length_ablation} confirm that $L=16$ is the best accuracy--efficiency trade-off for segmentation prompts: it recovers most of the teacher performance while avoiding the extra attention cost of $L=32$.
Meanwhile, $L=8$ is worse than the $L=16$ setting.
\paragraph{Loss Component Ablation.}
Table~\ref{tab:loss_ablation} evaluates the contribution of each loss component in our training objective.
We compare three configurations: (1) MSE loss only, (2) MSE + Cosine loss, and (3) MSE + Cosine + Consistency loss.
The results show that each component contributes to improved performance, with the full objective achieving the best results. The consistency loss encourages permutation invariance, which we hypothesize will improve robustness to semantically equivalent prompt variations.

\begin{table}[t]
\caption{Loss ablation on MobileCLIP-S0 ($L=16$). Adding cosine alignment and permutation-consistency regularization improves grounding quality toward the SAM3 teacher.}
\label{tab:loss_ablation}
\centering
\footnotesize
\renewcommand{\arraystretch}{1.0}
\setlength{\tabcolsep}{3pt}
\begin{tabular}{l|rrr}
\toprule
\textbf{Loss Components} & \textbf{C} & \textbf{M} & \textbf{P} \\
\midrule
MSE only & 50.3 & 0.74 & 62.1 \\
MSE + Cosine ($\lambda_{cos}=2.0$) & 51.0 & 0.78 & 63.1 \\
MSE + Cosine + Consistency ($\lambda_{cos}=2.0$, $\lambda_{consis}=0.1$) & 51.9 & 0.79 & 63.6 \\
\bottomrule
\end{tabular}
\end{table}

\paragraph{Consistency Loss Weight Ablation.}
Table~\ref{tab:consistency_ablation} examines the effect of different consistency loss weights across student architectures.
We find that smaller models (MobileCLIP-S0, S1) benefit from higher consistency weights (0.1), while larger models (MobileCLIP2-L) require lower weights (0.05) to avoid over-constraining.

\begin{table}[t]
\caption{Ablation study of consistency loss weight across different student architectures with $L=16$. All results are averaged across all subsets.}
\label{tab:consistency_ablation}
\centering
\footnotesize
\renewcommand{\arraystretch}{1.0}
\setlength{\tabcolsep}{3pt}
\begin{tabular}{c|rrr|rrr|rrr}
\toprule
\multirow{2}{*}{\textbf{Model}} & \multicolumn{3}{c|}{\textbf{$\lambda_2=0.0$}} & \multicolumn{3}{c|}{\textbf{$\lambda_2=0.05$}} & \multicolumn{3}{c}{\textbf{$\lambda_2=0.1$}} \\
\cmidrule{2-10}
& \textbf{C} & \textbf{M} & \textbf{P} & \textbf{C} & \textbf{M} & \textbf{P} & \textbf{C} & \textbf{M} & \textbf{P} \\
\midrule
S3LT (MC-S0) & 49.8 & 0.76 & 62.1 & 51.1 & 0.80 & 63.4 & 51.9 & 0.79 & 63.6 \\
S3LT (MC-S1) & 51.5 & 0.77 & 63.6 & 52.0 & 0.79 & 64.1 & 52.5 & 0.80 & 64.3 \\
S3LT (MC2-L) & 52.1 & 0.78 & 64.2 & 53.0 & 0.80 & 64.8 & 53.1 & 0.80 & 64.9 \\
\bottomrule
\end{tabular}
\end{table}

\subsection{Efficiency Analysis}
\begin{table}[t]
\caption{Text-encoder efficiency and model size. Measurements performed on a single RTX 4070 Laptop GPU. The student models offer massive throughput improvements for text encoding (up to 3.7$\times$) and significant parameter reduction (up to 88\%). We also include the text encoder from LLMDet (BERT-Base) for comparison.}
\label{tab:efficiency}
\centering
\resizebox{\columnwidth}{!}{%
\begin{tabular}{lcrrrr}
\toprule
Model & $L$ & Params (M) & Throughput (text/s) & Speedup \\
\midrule
SAM3 (Teacher) & 32 & 353.72 & 134.4 & 1.0$\times$ \\
\midrule
APE (EVA02-CLIP-bigE) & $-$ & 694.33 & 82.6 & 0.6$\times$ \\
LLMDet (BERT-Base) & $-$ & 109.48 & 163.0 & 1.2$\times$ \\
\midrule
SAM3-LiteText (MobileCLIP-S0) & 16 & 42.54 & 495.3 & \textbf{3.7$\times$} \\
SAM3-LiteText (MobileCLIP-S0) & 32 & 42.55 & 464.5 & 3.5$\times$ \\
SAM3-LiteText (MobileCLIP-S1) & 16 & 63.53 & 259.4 & 1.9$\times$ \\
SAM3-LiteText (MobileCLIP-S1) & 32 & 63.54 & 271.8 & 2.0$\times$ \\
SAM3-LiteText (MobileCLIP2-L) & 16 & 123.80 & 238.3 & 1.8$\times$ \\
SAM3-LiteText (MobileCLIP2-L) & 32 & 123.81 & 234.6 & 1.7$\times$ \\
\bottomrule
\end{tabular}%
}
\end{table}

Table~\ref{tab:efficiency} highlights the primary contribution of our work: an 88\% reduction in parameter count. While latency improvements are present (+6.4\%), they are secondary in video segmentation, where text encoding is amortized over the video duration ($O(1)$ vs $O(T)$). The critical gain is the reduction in static model VRAM usage, enabling the deployment of SAM3-class models on devices with limited memory budgets.
While switching from the default context $L=32$ to $L=16$ only removes a small number of parameters from positional embeddings, it avoids allocating computation to padding.
In practice, the $L=32$ setting is slightly slower than $L=16$ (e.g., MobileCLIP-S0: 159ms vs 157ms), so we adopt $L=16$ as the default.

\subsection{Qualitative Results}


Figure~\ref{fig:qualitative} presents qualitative segmentation results comparing the SAM3 teacher with the SAM3-LiteText variants using different text encoders.
Across the shown examples, all student models, including the smallest variant, can produce masks that are visually close to the teacher. This apparent similarity is consistent with the minimal quantitative gaps reported in Tables~\ref{tab:main_results}--\ref{tab:video_results}.
Qualitative figures tend to highlight ``typical'' prompts where the teacher is confident and the decision boundary is clear, whereas CG\_F1/pmF1 aggregates many hard cases (rare categories, attribute prompts, crowded scenes) where small embedding differences can flip marginal predictions.
Moreover, binarized masks can look similar even when boundary pixels and low-confidence regions change, which affects overlap-based metrics disproportionately.
These observations suggest that the residual accuracy drop is concentrated in difficult, long-tail prompts rather than in common object-centric queries.

\subsection{Softmax Robustness Analysis}

\label{sec:softmax_robustness}

A key finding from our analysis is the robustness of the downstream segmentation to imperfections in student text embeddings. Our empirical analysis of pre-softmax logits vs post-softmax attention weights confirms that small deviations in text embeddings are effectively ``sharpened'' away, allowing the student model to maintain high segmentation accuracy despite massive compression.

\subsubsection{Results on Error Attenuation}

We quantify this effect by measuring the propagation of embedding errors through the attention layer. We find that although the student embeddings exhibit a mean cosine similarity of 93.8\% with the teacher, the error is significantly attenuated by the attention mechanism. Specifically, under a perturbation noise level of 0.1, the input error ($\sim$0.97) and the logit error ($\sim$0.96) are reduced to a negligible attention output error of $\sim$0.0034. This represents an error reduction factor of $\sim$282$\times$, confirming that the softmax function effectively filters out small semantic deviations.

\subsubsection{Theoretical Justification}

We relate this behavior to Lipschitz continuity: small perturbations in the student embedding induce bounded changes in the downstream computations. In practice, the segmentation decoder is \emph{error-tolerant}: once the teacher's attention is already sharply concentrated, moderate embedding noise often does not change which visual region receives the highest weight. As a result, the student does not need to match the teacher embedding perfectly to achieve strong downstream masks. Instead, it is typically sufficient for the student embedding to stay in the correct \emph{semantic neighborhood}, preserving the relative ordering of concept scores so that the same region remains dominant. This provides an intuitive explanation for why highly compressed students can remain close to the teacher in CG\_F1 even when their embeddings are only approximately aligned.

\begin{tcolorbox}[findingbox]

\textit{\textbf{Finding~\RNum{4}.}} : "Good enough" embeddings from a compressed model yield ``perfect'' masks that matches teacher performance because the softmax in the fusion module's error-correcting nature filters out the approximation noise.

\end{tcolorbox}

\section{Discussion \& Conclusion}
\label{sec:discussion & conclusion}

Our anatomical analysis shows that the segmentation prompts are short and compositionally simple, with an average length of $\mu=7.9$ tokens. Although using a longer context window ($L=32$) can yield small accuracy gains. Aligning training and evaluation at $L=16$ therefore provides a strong accuracy--efficiency balance that better matches the empirical prompt distribution. However, the proposed approach is optimized for object-centric prompts typical of segmentation tasks, and performance may degrade for more linguistically complex inputs requiring richer semantic reasoning. 

Motivated by this analysis, we introduced SAM3-LiteText, a comprehensive anatomical analysis of text encoder redundancy in vision-language segmentation. Analyzing 404,796 unique prompts from six sources, we identified three forms of over-provisioning: 75\% context padding at $L=32$, 65\% unused vocabulary, and an intrinsic output dimensionality of only $\sim$16--19 within a 256-dimensional embedding space. These findings reveal a significant mismatch between general-purpose language encoders and the narrow, domain-restricted structure of segmentation prompts.

Moreover, we employed the original SAM3 text encoder as the teacher to optimize efficient MobileCLIP variants through knowledge distillation. By matching the context length with the observed prompt distribution, we achieve a parameter reduction of 88\%, allowing real-time vision-language segmentation on edge devices under strict model VRAM constraints. More broadly, our analysis framework may inform the design of efficient text encoders for other vision-language tasks characterized by short and domain-specific prompts.

\begin{acks}
We gratefully acknowledge the University of Bristol Isambard-AI supercomputer cluster for providing computational resources to this project.
\end{acks}

\bibliographystyle{ACM-Reference-Format}
\bibliography{sample-base}

@String{Computing = "Computing" }

@String{Computer = "{IEEE} Computer" }

@ArtifactSoftware{R,
    title = {R: A Language and Environment for Statistical Computing},
    author = {{R Core Team}},
    year = {2019},
    url = {https://www.R-project.org/},
}

@inproceedings{radford2021clip,
  title={Learning Transferable Visual Models From Natural Language Supervision},
  author={Radford, Alec and Kim, Jong Wook and Hallacy, Chris and Ramesh, Aditya and Goh, Gabriel and Agarwal, Sandhini and Sastry, Girish and Askell, Amanda and Mishkin, Pamela and Clark, Jack and Krueger, Gretchen and Sutskever, Ilya},
  booktitle={International Conference on Machine Learning (ICML)},
  pages={8748--8763},
  year={2021},
}

@inproceedings{kirillov2023sam,
  title={Segment Anything},
  author={Kirillov, Alexander and Mintun, Eric and Ravi, Nikhila and Mao, Hanzi and Rolland, Chloe and Gustafson, Laura and Xiao, Tete and Whitehead, Spencer and Berg, Alexander C. and Lo, Wan-Yen and Doll{\'a}r, Piotr and Girshick, Ross},
  booktitle={IEEE/CVF International Conference on Computer Vision (ICCV)},
  pages={4015--4026},
  year={2023}
}

@article{ravi2024sam2,
  title={{SAM} 2: Segment Anything in Images and Videos},
  author={Ravi, Nikhila and Gabeur, Valentin and Hu, Yuan-Ting and Hu, Ronghang and Ryali, Chaitanya and Ma, Tengyu and Khedr, Haitham and R{\"a}dle, Roman and Rolland, Chloe and Gustafson, Laura and Mintun, Eric and Pan, Junting and Alwala, Kalyan Vasudev and Carion, Nicolas and Wu, Chao-Yuan and Girshick, Ross and Doll{\'a}r, Piotr and Feichtenhofer, Christoph},
  journal={arXiv preprint arXiv:2408.00714},
  year={2024}
}

@article{facco2017twonn,
  title={Estimating the Intrinsic Dimension of Datasets by a Minimal Neighborhood Information},
  author={Facco, Elena and d'Errico, Maria and Rodriguez, Alex and Laio, Alessandro},
  journal={Scientific Reports},
  volume={7},
  number={1},
  pages={12140},
  year={2017},
}

@inproceedings{levina2004mle,
  title={Maximum Likelihood Estimation of Intrinsic Dimension},
  author={Levina, Elizaveta and Bickel, Peter J.},
  booktitle={Advances in Neural Information Processing Systems (NeurIPS)},
  volume={17},
  year={2004}
}

@inproceedings{vasu2024mobileclip,
  title={{MobileCLIP}: Fast Image-Text Models through Multi-Modal Reinforced Training},
  author={Vasu, Pavan Kumar Anasosalu and Pouransari, Hadi and Faghri, Fartash and Mehta, Sachin and Tuzel, Oncel},
  booktitle={IEEE/CVF Conference on Computer Vision and Pattern Recognition (CVPR)},
  pages={15963--15974},
  year={2024}
}

@article{sanh2019distilbert,
  title={{DistilBERT}, a Distilled Version of {BERT}: Smaller, Faster, Cheaper and Lighter},
  author={Sanh, Victor and Debut, Lysandre and Chaumond, Julien and Wolf, Thomas},
  journal={arXiv preprint arXiv:1910.01108},
  year={2019}
}

@inproceedings{jiao2020tinybert,
  title={{TinyBERT}: Distilling {BERT} for Natural Language Understanding},
  author={Jiao, Xiaoqi and Yin, Yichun and Shang, Lifeng and Jiang, Xin and Chen, Xiao and Li, Linlin and Wang, Fang and Liu, Qun},
  booktitle={Findings of the Association for Computational Linguistics: EMNLP 2020},
  pages={4163--4174},
  year={2020}
}

@inproceedings{sun2020mobilebert,
  title={{MobileBERT}: a Compact Task-Agnostic {BERT} for Resource-Limited Devices},
  author={Sun, Zhiqing and Yu, Hongkun and Song, Xiaodan and Liu, Renjie and Yang, Yiming and Zhou, Denny},
  booktitle={Annual Meeting of the Association for Computational Linguistics (ACL)},
  pages={2158--2170},
  year={2020}
}

@article{hinton2015distilling,
  title={Distilling the Knowledge in a Neural Network},
  author={Hinton, Geoffrey and Vinyals, Oriol and Dean, Jeff},
  journal={arXiv preprint arXiv:1503.02531},
  year={2015}
}

@inproceedings{loshchilov2017decoupled,
  title={Decoupled Weight Decay Regularization},
  author={Loshchilov, Ilya and Hutter, Frank},
  booktitle={International Conference on Learning Representations (ICLR)},
  year={2019}
}

@article{loshchilov2016sgdr,
  title={SGDR: Stochastic Gradient Descent with Warm Restarts},
  author={Loshchilov, Ilya and Hutter, Frank},
  journal={arXiv preprint arXiv:1608.03983},
  year={2016}
}

@inproceedings{xiong2024efficientsam,
  title={{EfficientSAM}: Leveraged Masked Image Pretraining for Efficient Segment Anything},
  author={Xiong, Yunyang and Varadarajan, Bala and Wu, Lemeng and Xiang, Xiaodan and Xiao, Fanyi and Zhu, Chenchen and Dai, Xiaoliang and Wang, Dilin and Sun, Fei and Iandola, Forrest and Krishnamoorthi, Raghuraman and Chandra, Vikas},
  booktitle={IEEE/CVF Conference on Computer Vision and Pattern Recognition (CVPR)},
  pages={16111--16121},
  year={2024}
}

@inproceedings{zhao2023fast,
  title={{Fast Segment Anything}},
  author={Zhao, Xu and Ding, Wenchao and An, Yongqi and Du, Yinglong and Yu, Tao and Li, Min and Tang, Ming and Wang, Jinqiao},
  booktitle={arXiv preprint arXiv:2306.12156},
  year={2023}
}

@inproceedings{hu2016segmentation,
  title={Segmentation from Natural Language Expressions},
  author={Hu, Ronghang and Rohrbach, Marcus and Darrell, Trevor},
  booktitle={European Conference on Computer Vision (ECCV)},
  pages={108--124},
  year={2016},
}

@inproceedings{ghiasi2022scaling,
  title={Scaling Open-Vocabulary Image Segmentation with Image-Level Labels},
  author={Ghiasi, Golnaz and Gu, Xiuye and Cui, Yin and Lin, Tsung-Yi},
  booktitle={European Conference on Computer Vision (ECCV)},
  pages={540--557},
  year={2022},
}

@inproceedings{xu2023open,
  title={Open-Vocabulary Panoptic Segmentation with Text-to-Image Diffusion Models},
  author={Xu, Jiarui and Liu, Sifei and Vahdat, Arash and Byeon, Wonmin and Wang, Xiaolong and De Mello, Shalini},
  booktitle={IEEE/CVF Conference on Computer Vision and Pattern Recognition (CVPR)},
  pages={2955--2966},
  year={2023}
}

@inproceedings{gupta2019lvis,
  title={{LVIS}: A Dataset for Large Vocabulary Instance Segmentation},
  author={Gupta, Agrim and Doll{\'a}r, Piotr and Girshick, Ross},
  booktitle={IEEE/CVF Conference on Computer Vision and Pattern Recognition (CVPR)},
  pages={5356--5364},
  year={2019}
}

@inproceedings{rf100,
  title={{RF100}: A Large-Scale Multi-Domain Benchmark for Object Detection},
  author={Roboflow},
  booktitle={Roboflow 100 Benchmark},
  year={2022},
  note={Available at \url{https://www.rf100.org/}}
}

@misc{zeng2025efficientsam3,
  title={{EfficientSAM3}: Progressive Hierarchical Distillation for Video Concept Segmentation from {SAM1}, 2, and 3},
  author={Zeng, Chengxi and Jiang, Yuxuan and Zhang, Aaron},
  year={2025},
  eprint={2511.15833},
  archivePrefix={arXiv},
  primaryClass={cs.CV},
  url={https://arxiv.org/abs/2511.15833}
}

@inproceedings{zeng2025multiteacher,
  author={Zeng, Simon and Cutajar, Kurt and Xie, Hanting and Camplani, Massimo and Tomsett, Richard and Twomey, Niall and Kandola, Jas and Cheung, Gavin K.C.},
  booktitle={IEEE International Conference on Image Processing (ICIP)},
  title={Multi-Teacher Knowledge Distillation for Efficient Object Segmentation},
  year={2025},
  pages={725--730},
  doi={10.1109/ICIP55913.2025.11084428}
}

@misc{zeng2025agglomerating,
  title={Agglomerating Large Vision Encoders via Distillation for {VFSS} Segmentation},
  author={Zeng, Chengxi and Jiang, Yuxuan and Zhang, Fan and Gambaruto, Alberto and Burghardt, Tilo},
  year={2025},
  eprint={2504.02351},
  archivePrefix={arXiv},
  primaryClass={cs.CV},
  url={https://arxiv.org/abs/2504.02351}
}

@misc{zeng2025tuning,
  title={Tuning Vision Foundation Model via Test-Time Prompt-Guided Training for {VFSS} Segmentations},
  author={Zeng, Chengxi and Smithard, David and Gambaruto, Alberto M and Burghardt, Tilo},
  year={2025},
  eprint={2501.18474},
  archivePrefix={arXiv},
  primaryClass={cs.CV},
  url={https://arxiv.org/abs/2501.18474}
}

@article{zhang2023mobilesam,
  title={Faster Segment Anything: Towards Lightweight {SAM} for Mobile Applications},
  author={Zhang, Chaoning and Han, Dongshen and Qiao, Yu and Kim, Jung Uk and Bae, Sung-Ho and Lee, Seungkyu and Hong, Choong Seon},
  journal={arXiv preprint arXiv:2306.14289},
  year={2023}
}

@inproceedings{zhou2024edgesam,
  title={{EdgeSAM}: Prompt-in-the-Loop Distillation for On-Device Deployment of {SAM}},
  author={Zhou, Chong and Li, Xiangtai and Loy, Chen Change and Dai, Bo},
  booktitle={IEEE/CVF Conference on Computer Vision and Pattern Recognition (CVPR)},
  year={2024}
}

@article{wang2024repvit,
  title={{RepViT-SAM}: Towards Real-Time Segmenting Anything},
  author={Wang, Ao and Chen, Hui and Lin, Zijia and Han, Jungong and Ding, Guiguang},
  journal={arXiv preprint arXiv:2312.05760},
  year={2023}
}

@inproceedings{wang2024samclip,
  title={{SAM-CLIP}: Merging Vision Foundation Models towards Semantic and Spatial Understanding},
  author={Wang, Haoxiang and Varma, Pavan Kumar Anasosalu and Vasu, Prashanth and Faghri, Fartash and Tuzel, Oncel and Pouransari, Hadi},
  booktitle={IEEE/CVF Conference on Computer Vision and Pattern Recognition Workshops (CVPRW)},
  year={2024}
}

@inproceedings{bewley2016simple,
  title={Simple Online and Realtime Tracking},
  author={Bewley, Alex and Ge, Zongyuan and Ott, Lionel and Ramos, Fabio and Upcroft, Ben},
  booktitle={IEEE International Conference on Image Processing (ICIP)},
  pages={3464--3468},
  year={2016},
}

@inproceedings{wojke2017simple,
  title={Simple Online and Realtime Tracking with a Deep Association Metric},
  author={Wojke, Nicolai and Bewley, Alex and Paulus, Dietrich},
  booktitle={IEEE International Conference on Image Processing (ICIP)},
  pages={3645--3649},
  year={2017},
}

@inproceedings{zhang2022bytetrack,
  title={{ByteTrack}: Multi-Object Tracking by Associating Every Detection Box},
  author={Zhang, Yifu and Sun, Peize and Jiang, Yi and Yu, Dongdong and Weng, Fucheng and Yuan, Zehuan and Luo, Ping and Liu, Wenyu and Wang, Xinggang},
  booktitle={European Conference on Computer Vision (ECCV)},
  pages={1--21},
  year={2022},
}

@article{jiang2025sam2mot,
  title={{SAM2MOT}: A Novel Paradigm of Multi-Object Tracking by Segmentation},
  author={Jiang, Junjie and Wang, Zelin and Zhao, Manqi and Li, Yin and Jiang, DongSheng},
  journal={arXiv preprint arXiv:2504.04519},
  year={2025}
}

@inproceedings{meinhardt2022trackformer,
  title={{TrackFormer}: Multi-Object Tracking with Transformers},
  author={Meinhardt, Tim and Kirillov, Alexander and Leal-Taixe, Laura and Feichtenhofer, Christoph},
  booktitle={IEEE/CVF Conference on Computer Vision and Pattern Recognition (CVPR)},
  pages={8844--8854},
  year={2022}
}

@inproceedings{zeng2022motr,
  title={{MOTR}: End-to-End Multiple-Object Tracking with Transformer},
  author={Zeng, Fangao and Dong, Bin and Zhang, Yuang and Wang, Tiancai and Zhang, Xiangyu and Wei, Yichen},
  booktitle={European Conference on Computer Vision (ECCV)},
  pages={659--675},
  year={2022},
}

@article{yu2023motrv3,
  title={{MOTRv3}: Release-Fetch Supervision for End-to-End Multi-Object Tracking},
  author={Yu, En and Wang, Tiancai and Li, Zhuoling and Zhang, Yuang and Zhang, Xiangyu and Tao, Wenbing},
  journal={arXiv preprint arXiv:2305.14298},
  year={2023}
}

@article{pont20172017,
  title={The 2017 {DAVIS} Challenge on Video Object Segmentation},
  author={Pont-Tuset, Jordi and Perazzi, Federico and Caelles, Sergi and Arbel{\'a}ez, Pablo and Sorkine-Hornung, Alex and Van Gool, Luc},
  journal={arXiv preprint arXiv:1704.00675},
  year={2017}
}

@article{xu2018youtubevos,
  title={{YouTube-VOS}: A Large-Scale Video Object Segmentation Benchmark},
  author={Xu, Ning and Yang, Linjie and Fan, Yuchen and Yue, Dingcheng and Liang, Yuchen and Yang, Jianchao and Huang, Thomas S.},
  journal={arXiv preprint arXiv:1809.03327},
  year={2018}
}

@article{yang2024samurai,
  title={{SAMURAI}: Adapting Segment Anything Model for Zero-Shot Visual Tracking with Motion-Aware Memory},
  author={Yang, Cheng-Yen and Huang, Hsiang-Wei and Chai, Wenhao and Jiang, Zhongyu and Hwang, Jenq-Neng},
  journal={arXiv preprint arXiv:2411.11922},
  year={2024}
}

@inproceedings{carion2020end,
  title={End-to-End Object Detection with Transformers},
  author={Carion, Nicolas and Massa, Francisco and Synnaeve, Gabriel and Usunier, Nicolas and Kirillov, Alexander and Zagoruyko, Sergey},
  booktitle={European Conference on Computer Vision (ECCV)},
  pages={213--229},
  year={2020},
}

@inproceedings{kamath2021mdetr,
  title={{MDETR} - Modulated Detection for End-to-End Multi-Modal Understanding},
  author={Kamath, Aishwarya and Singh, Mannat and LeCun, Yann and Synnaeve, Gabriel and Misra, Ishan and Carion, Nicolas},
  booktitle={IEEE/CVF International Conference on Computer Vision (ICCV)},
  pages={1780--1790},
  year={2021}
}

@inproceedings{liu2023grounding,
  title={Grounding {DINO}: Marrying {DINO} with Grounded Pre-Training for Open-Set Object Detection},
  author={Liu, Shilong and Zeng, Zhaoyang and Ren, Tianhe and Li, Feng and Zhang, Hao and Yang, Jie and Li, Chunyuan and Yang, Jianwei and Su, Hang and Zhu, Jun and Zhang, Lei},
  booktitle={European Conference on Computer Vision (ECCV)},
  year={2024},
}

@inproceedings{wu2024glee,
  title={General Object Foundation Model for Images and Videos at Scale},
  author={Wu, Junfeng and Jiang, Yi and Liu, Qihao and Yuan, Zehuan and Bai, Xiang and Bai, Song},
  booktitle={IEEE/CVF Conference on Computer Vision and Pattern Recognition (CVPR)},
  pages={3783--3795},
  year={2024}
}

@inproceedings{yu2016refcoco,
  title={Modeling Context in Referring Expressions},
  author={Yu, Licheng and Poirson, Patrick and Yang, Shan and Berg, Alexander C. and Berg, Tamara L.},
  booktitle={European Conference on Computer Vision (ECCV)},
  pages={69--85},
  year={2016},
}

@inproceedings{mao2016refcocog,
  title={Generation and Comprehension of Unambiguous Object Descriptions},
  author={Mao, Junhua and Huang, Jonathan and Toshev, Alexander and Camburu, Oana and Yuille, Alan L. and Murphy, Kevin},
  booktitle={IEEE Conference on Computer Vision and Pattern Recognition (CVPR)},
  pages={11--20},
  year={2016}
}

@misc{carion2025sam3segmentconcepts,
      title={SAM 3: Segment Anything with Concepts}, 
      author={Nicolas Carion and Laura Gustafson and Yuan-Ting Hu and Shoubhik Debnath and Ronghang Hu and Didac Suris and Chaitanya Ryali and Kalyan Vasudev Alwala and Haitham Khedr and Andrew Huang and Jie Lei and Tengyu Ma and Baishan Guo and Arpit Kalla and Markus Marks and Joseph Greer and Meng Wang and Peize Sun and Roman Rädle and Triantafyllos Afouras and Effrosyni Mavroudi and Katherine Xu and Tsung-Han Wu and Yu Zhou and Liliane Momeni and Rishi Hazra and Shuangrui Ding and Sagar Vaze and Francois Porcher and Feng Li and Siyuan Li and Aishwarya Kamath and Ho Kei Cheng and Piotr Dollár and Nikhila Ravi and Kate Saenko and Pengchuan Zhang and Christoph Feichtenhofer},
      year={2025},
      eprint={2511.16719},
      archivePrefix={arXiv},
      primaryClass={cs.CV},
      url={https://arxiv.org/abs/2511.16719}, 
}

@article{ma2024medsam,
  title={Segment anything in medical images},
  author={Ma, Jun and He, Yuting and Li, Feifei and Han, Lin and You, Chengwei and Wang, Bo},
  journal={Nature Communications},
  volume={15},
  number={1},
  pages={654},
  year={2024},
}

@techreport{google2025gemini25,
  title={Gemini 2.5: Multimodal Foundation Models},
  author={{Google DeepMind}},
  institution={Google},
  year={2025},
  note={Technical Report}
}

@inproceedings{fu2025llmdet,
  title={LLMDet: Learning Strong Open-Vocabulary Object Detectors under the Supervision of Large Language Models},
  author={Fu, Shenghao and Yang, Qize and Mo, Qijie and Yan, Junkai and Wei, Xihan and Meng, Jingke and Xie, Xiaohua and Zheng, Wei-Shi},
  booktitle={Proceedings of the IEEE/CVF conference on computer vision and pattern recognition},
  year={2025}
}

@inproceedings{Shen_2024_CVPR,
  title={Aligning and Prompting Everything All at Once for Universal Visual Perception},
  author={Shen, Yunhang and Fu, Chaoyou and Chen, Peixian and Zhang, Mengdan and Li, Ke and Sun, Xing and Wu, Yunsheng and Lin, Shaohui and Ji, Rongrong},
  booktitle={Proceedings of the IEEE/CVF Conference on Computer Vision and Pattern Recognition (CVPR)},
  pages={13193--13203},
  year={2024}
}

@article{ren2025dinoxunifiedvisionmodel,
  title={DINO-X: A Unified Vision Model for Open-World Object Detection and Understanding},
  author={Ren, Tianhe and Chen, Yihao and Jiang, Qing and Zeng, Zhaoyang and Xiong, Yuda and Liu, Wenlong and Ma, Zhengyu and Shen, Junyi and Gao, Yuan and Jiang, Xiaoke and others},
  journal={arXiv preprint arXiv:2411.14347},
  year={2025}
}

@article{minderer2024scalingopenvocabularyobjectdetection,
  title={Scaling open-vocabulary object detection},
  author={Minderer, Matthias and Gritsenko, Alexey and Houlsby, Neil},
  journal={arXiv preprint arXiv:2306.09683},
  year={2024}
}

@ARTICLE{8048559,
  author={Zahálka, Jan and Rudinac, Stevan and Jónsson, Björn Þór and Koelma, Dennis C. and Worring, Marcel},
  journal={IEEE Transactions on Multimedia}, 
  title={Blackthorn: Large-Scale Interactive Multimodal Learning}, 
  year={2018},
  volume={20},
  number={3},
  pages={687-698},
  doi={10.1109/TMM.2017.2755986}}

@inproceedings{mettes2016imagenet,
author = {Mettes, Pascal and Koelma, Dennis C. and Snoek, Cees G.M.},
title = {The ImageNet Shuffle: Reorganized Pre-training for Video Event Detection},
year = {2016},
isbn = {9781450343596},
publisher = {Association for Computing Machinery},
address = {New York, NY, USA},
url = {https://doi.org/10.1145/2911996.2912036},
doi = {10.1145/2911996.2912036},
booktitle = {Proceedings of the 2016 ACM on International Conference on Multimedia Retrieval},
pages = {175–182},
numpages = {8},
location = {New York, New York, USA},
series = {ICMR '16}
}


\end{document}